\documentclass[10pt,twocolumn,letterpaper]{article}
\usepackage{balance}
\usepackage{wacv}
\usepackage{epsfig}
\usepackage{graphicx}
\usepackage{amsmath}
\usepackage{amssymb}
\usepackage{times}
\usepackage{soul}
\usepackage{url}
\usepackage[utf8]{inputenc}
\usepackage[small]{caption}
\usepackage{amsmath}
\usepackage{booktabs}
\usepackage{algorithm}
\usepackage{marginnote}
\usepackage[export]{adjustbox}
\usepackage{rotating}
\usepackage{calrsfs}
\usepackage{amsfonts}
\usepackage{siunitx}
\usepackage{tabularx}
\usepackage{hhline}
\usepackage{subcaption}
\usepackage[noend]{algpseudocode}
\usepackage{multirow}
\usepackage{makecell}
\usepackage{textcomp}
\DeclareMathAlphabet{\pazocal}{OMS}{zplm}{m}{n}

\newcommand{\degree}[1]{${#1}^o$}
\def\360{\degree{360}}
\newcommand{\srcTarget}[3]{$\text{#1}(\text{#2}) \rightarrow \text{#3}$}
\usepackage[pagebackref=true,breaklinks=true,letterpaper=true,colorlinks,bookmarks=false]{hyperref}

\begin{document}

\wacvfinalcopy
%%%%%%%%% TITLE
\title{Restyling Data: Application to Unsupervised Domain Adaptation}

\author{Vasileios Gkitsas
\quad \quad \quad \quad 
Antonis Karakottas
\quad \quad \quad \quad 
Nikolaos Zioulis
\quad \quad
\\
Dimitrios Zarpalas
\quad \quad \quad
Petros Daras
\\
{\tt\small{Information Technologies Institute (ITI), Centre for Research and Technology Hellas (CERTH), Greece} }\\
{\tt\small{\{gkitsasv, ankarako, nzioulis, zarpalas, daras\}@iti.gr}}\\
\\ 
\vspace*{-1cm}}
%{\tt\small {
%gkitsasv, ankarako, nzioulis, zarpalas, daras}@iti.gr}

\maketitle
\thispagestyle{empty}
\begin{abstract}
%Our method is a natural and intuitive application of FashPhotoStyle \cite{fastphoto} to domain adaptation
Machine learning is driven by data, yet while their availability is constantly increasing, training data require laborious, time consuming and error-prone labelling or ground truth acquisition, which in some cases is very difficult or even impossible. 
Recent works have resorted to synthetic data generation, but the inferior performance of models trained on synthetic data when applied to the real world, introduced the challenge of unsupervised domain adaptation.
In this work we investigate an unsupervised domain adaptation technique that descends from another perspective, in order to avoid the complexity of adversarial training and cycle consistencies. 
We exploit the recent advances in photorealistic style transfer and take a fully data driven approach. 
While this concept is already implicitly formulated within the intricate objectives of domain adaptation GANs, we take an explicit approach and apply it directly as data pre-processing.
The resulting technique is scalable, efficient and easy to implement, offers competitive performance to the complex state-of-the-art alternatives and can open up new pathways for domain adaptation. 
%We showcase competitive performance in a variety of domain adaptation tasks and datasets.
\end{abstract}
%%%%%%%%% BODY TEXT
\section{Introduction}
\label{sec:intro}
The proven capability of deep convolutional neural networks (CNNs) to represent complex functions has been steadily pushing their adoption in many computer vision tasks.
A great deal of work focuses on boosting their performance by increasing their expressiveness through architectural modifications or richer supervision schemes.
Nonetheless, for most vision related tasks, the main driving force behind deep CNNs is unarguably the amount of training data. 

From a practical standpoint, their real world applicability is related to their efficacy when applied to in-the-wild settings.
In other words, each CNN's capability to generalize and retain its performance in a variety of environments and conditions, is of paramount importance.
There are many techniques that are being employed to address this, like dropping neurons \cite{srivastava2014dropout} or layers \cite{huang2016deep}, parameter sharing and regularization, all being different ways to prevent over-fitting.
However, increasing the amount and diversity of the training data is also an indirect way to increase their generalization capacity and prevent over-fitting.

%\vspace{-2pt}
\begin{figure}[!t]
    \definecolor{awesome}{rgb}{1.0, 0.13, 0.32}
    \definecolor{amethyst}{rgb}{0.6, 0.4, 0.8}
    \definecolor{blue-green}{rgb}{0.0, 0.87, 0.87}
    \definecolor{azure}{rgb}{0.0, 0.5, 1.0}
    \centering
    \includegraphics[width=\columnwidth]{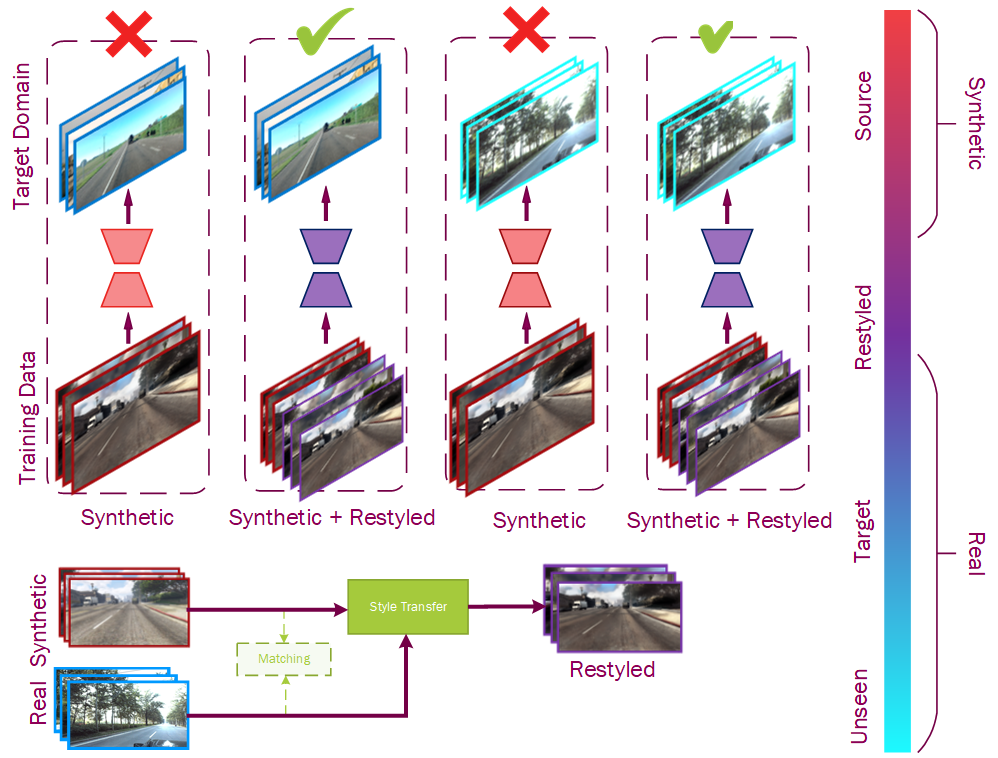}
    \caption{Overview of the proposed method. We use style transfer explicitly to align data distributions at the dataset level.
    Starting from a \textcolor{awesome}{labelled synthetic source} domain we select samples from an \textcolor{azure}{unlabelled realistic target} domain and restyle them to create a \textcolor{amethyst}{labelled restyled} domain, more aligned to the target one. This is used to enrich the original labelled dataset.
    Data selection can be either random or matched via perceptual hashing (Sec.~\ref{sec:approach}).
    The enriched dataset is used to train a model that learns feature representations more aligned to the realistic target domain used to restyle the original synthetic source data.
    This improves its performance when applied to data from the unlabelled target domain, compared to training only with the labelled source data.
    Further, the model shows performance gains even when applied to \textcolor{blue-green}{other realistic domains (unseen)} which are closer to the target domain than the source (as indicated by the domain alignment color scale on the right). Best viewed in color. 
    %We consider four domains, namely {\color{awesome} source}, {\color{amethyst} restyled}, {\color{azure} target} and {\color{blue-green} unseen}, from which we consider that the {\color{awesome} source} domain is composed of synthetic data, and both the {\color{azure} target} and {\color{blue-green} unseen} domains are subsets of real data. We employ style transfer by matching real and synthetic data, either randomly (RS) or by perceptual hashing (PH) (described in Section \ref{sec:approach}) to bridge the gap between the synthetic and real, from which we produce restyled data. First, we train a CNN to learn a vision task with only the synthetic data and evaluate it on the real domain, which produces subpar results. We then use both the synthetic and restyled data for training, which increases the performance of the learned model. Finally, we evaluate both models on real but unseen data. Once more, the model trained with synthetic data produces inferior results, while the one trained with both the synthetic and the restyled data increases performance by a limited amount.
    }
    %\caption{Overview of the proposed method. Training set consists of two distributions, synthetic $X_{s}$ and restyled $X_{re}$. The restyled distribution is produced via style transfer \cite{fastphoto}, using either random style selection or a matching selection strategy. (see section \ref{approach})}
    \label{fig:system_overview}
\end{figure}

%Given that labelled data acquisition is a costly and inefficient process, their automatic generation is a promising alternative.
%The synthesis of photo-realistic content by leveraging computer graphics technologies, has enabled automatic data generation \cite{gaidon2016virtual}.
\pagebreak
As labelled data acquisition is costly and inefficient, photo-realistic automatic data synthesis via computer graphics technologies is a promising alternative \cite{gaidon2016virtual}, with another one being data recording from computer games \cite{GTA}.
%Another interesting alternative is the recording of data from computer games \cite{GTA}, an approach that can also automatically label them by exploiting the assets' meta-data.
Nevertheless, besides the inherent dataset bias \cite{torralba2011unbiased}, such automatic data generation methods suffer from the synthetic-to-real (S2R) domain discrepancy \cite{yosinski2014transferable} that deteriorates the performance of models trained using rendered content when applied to real world acquired inputs.

State-of-the-art S2R domain adaptation methods typically utilize Generative Adversarial Networks (GANs) \cite{goodfellow2014generative} to align the source (synthetic) and target (real) domains under an adversarial framework.
Different variants exist that operate either in the image \cite{bousmalis2017unsupervised, liu2016coupled} or feature space \cite{ganin2015unsupervised}.
However, these approaches do not generalize well to other tasks \cite{zhang2017curriculum}, leading to the development of task specific approaches \cite{sankaranarayanan2018learning}.
Other approaches rely on image-to-image translation to create a mapping between the source and target domains \cite{stein2018genesis} or employ GANs to transfer the style of the target domain to the labelled source images \cite{atapour2018real}.

\begin{figure}[!ht]
    \centering
  
    \includegraphics[width=\columnwidth]{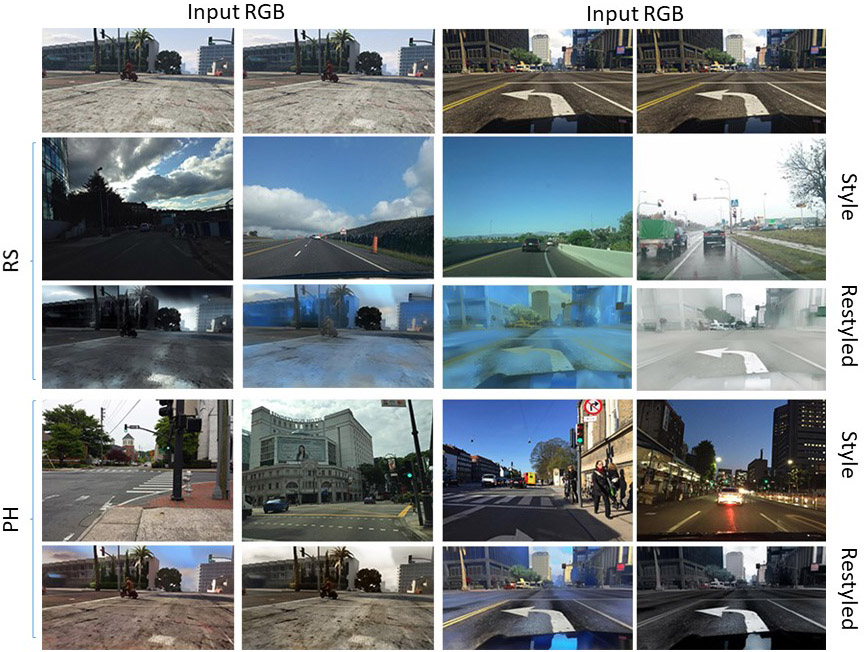}
    \label{subfig:bad}

%    \begin{subfigure}[b]{1.0\columnwidth}
%        \includegraphics[width=\columnwidth]{figures/img/good_bad_restyles/ph_selections.jpg}
%        \subcaption{}
%        \label{subfig:good}
%    \end{subfigure}
    
    \label{fig:goodBadRs}
    \caption{Examples showcasing the visual quality increase when matching styles randomly (RS), compared to style selection via perceptual hashing (PH). For each subfigure from top to bottom: the source image (top), the selected style (middle) and the restyled result (bottom). Uncanny effects are observed when randomly picking out the styles, compared to when using PH as the style-matching process, which improves the overall quality of the result.}
    \label{fig:goodbad}
\end{figure}

GAN based approaches though, require the training of a generator-discriminator pair which is a challenging task. 
They also  reduce the effective expressiveness and training time of the underlying model that is trained for the targeted task, as the adaptation itself consumes model capacity and training resources.
In actual industrial or production settings, new and diverse data can be acquired at enormous rates, but labelling them is a tedious and labour intensive task.
The suitability of GAN based solutions in production settings is questionable, as their associated challenges would be further magnified.
Therefore, novel solutions are required that are scalable and flexible. 

In this work, we explore one such alternative and find it to be very promising due to its applicability.
We showcase how the recent advances in direct style transfer \cite{Gatys} can bridge the domain gap, even when applied purely as data pre-processing.
More importantly, our approach operates in an unsupervised manner as it does not require cross domain labels and, accordingly, it can readily use an abundance of unlabelled data.
%This approach is straightforward and effective, as well as generic across different types of data and tasks.
%While similar archival works have recently surfaced \cite{jackson2018style, dundar208domain}, our contributions go beyond by:
Our concept is illustratively presented in Figure ~\ref{fig:system_overview}.
In summary our contributions are: \textbf{i)} We demonstrate a straightforward and effective technique for unsupervised domain adaptation (UDA) that offers competitive performance against the state-of-the-art, but without its complexity. \textbf{ii)} We present a sample selection process that is suitable for appropriately matching labelled samples of the source data distribution with unlabelled ones from the target data distribution, in order to improve the performance of our proposed technique. \textbf{iii)} We extensively evaluate our approach and demonstrate its generality across different tasks and datasets. \textbf{iv)} Unlike most similar approaches, we take a step beyond and consider \textit{target} domains that - albeit contextually similar - are different from both the labelled (\textit{source}), as well as the unlabelled (\textit{style}) domains.
\section{Related work}
\label{sec:related}

UDA aims at overcoming the domain shift between a \textit{source} domain used to train a machine learning model, and the \textit{target} domain that it will be applied to, whose data distribution differs from the source one. 
The lack of supervision implies that labelled data are only available for the source domain. 
Vision related machine learning tasks mostly rely on CNNs, and consequently, preliminary works introduced domain classifiers along with gradient reversal on their results, adding another objective for the CNN to optimize for, which ensured domain invariant feature learning \cite{ganin2015unsupervised}. 
Initial results were presented for digit \cite{MNIST} and object \cite{saenko2010adapting} classification, and later extended to other tasks \cite{GaninJournal}.
However, the datasets used are limited by today's standards in terms of scale and domain gap benchmarking potential.

With the introduction of the highly performing GANs \cite{goodfellow2014generative}, most of the recent UDA methods rely on adversarial objectives between the source and target data distributions.
A two-fold approach was presented in \cite{inthewild} with global domain alignment enforced by a domain classifier, operating on the encoded representations of the source and target domains, and local domain adaptation enforced by optimizing for consistency between the category probabilities of each distribution. As this approach is driven by the assumption that both domains contain similar distribu-
tions of the categories, it is limited only to dense segmentation scenarios
%As this approach is driven by the assumption that both domains contain similar distributions of %the categories, it is limited only to dense segmentation scenarios. 
While quantitative results were presented using SYNTHIA \cite{SYNTHIA} and GTA \cite{GTA} as source domains and Cityscapes (CS) \cite{cityscapes} as the target domain, an in-the-wild dataset was also collected for assessing adaptation between real world environments as well. 
In \cite{zhang2017curriculum}, focusing on semantic segmentation, a CNN is trained with direct supervision on the source domain, as well as properties consistent with the target domain like global and local label distributions, with the latter enforced via superpixel landmarks. Results were presented for SYN-
THIA and Cityscapes as the source and target domains respectively.
%Results were presented for SYNTHIA and Cityscapes as the source and target domains respectively.  

Other works \cite{im2im2,CyCADA} approached the problem of UDA under an unpaired image-to-image translation framework. 
In the former \cite{im2im2}, a complex optimization was formulated using numerous objectives that are weighted together to learn a domain agnostic feature representation for both the source and the target distributions. Overall, inter domain translations were used, along with cycle consistencies between them and translation classification losses.
%Overall, inter domain translations were used, along with cycle consistencies between them and %translation classification losses. 
In the latter \cite{CyCADA}, the concept of cycle consistent adversarial networks \cite{im2imCycle} is used to produce mappings between the source and target domains. 
Reconstruction and discriminator losses were utilized for the $source \rightarrow target \rightarrow source$ and the $source \rightarrow target$ mappings respectively, in addition to a task specific loss for the second mapping. 
As both of these frameworks are generic, they were applied to the digit classification and semantic segmentation tasks.

SG-GAN \cite{semAwareGradGAN} complements cycle consistencies and adversarial discriminators with a soft-gradient objective that aims to align texture edges with semantic boundaries, and presents results for the GTA to CS semantic segmentation task. 
Similarly, \cite{sankaranarayanan2018learning} utilizes direct supervision on the source domain, shared weights auto-encoders with an adversarial loss, to not only reconstruct but also distinguish the domains between the reconstructions, as well as an auxiliary supervision on the reconstructed image. Their provided results focus on virtual to real adaptation using the SYNTHIA and GTA datasets that are evaluated on Cityscapes.
%Their provided results focus on virtual to real adaptation using the SYNTHIA and GTA datasets %that are evaluated on Cityscapes.
Taking a slightly more original approach, \cite{Tsai_2018_CVPR} chose to discriminate based on the output semantic segmentation map on multiple scales and demonstrate their method under a S2R theme using GTA and SYNTHIA adapted to CS, and also under a real-to-real scheme, using CS and a Cross-City dataset \cite{chen2017no}. 

Perhaps one of the more original ideas is based on the adversarial dropout regularization concept \cite{dropoutRegularization} that searches for non-discriminative features on the shared encoding, and aims to achieve a better separation of the distribution at their boundaries.

More recently, the issues that traditional batch normalization creates during domain adaptation were acknowledged by \cite{romijnders2018domain}, and then adapted this technique appropriately for domain adaptation.
A two-step training process was used that involved first training a domain discriminator, and then using it to drive a confusion loss while directly supervising the source domain.
Besides offering results for the GTA to CS task, they also reported performance on unseen real data, and more specifically, in Mapillary Vistas (Map) \cite{mapillary} and ApolloScape (AS) \cite{apollo}.
In \cite{DAclassBalanced} the novel concept of self-training is introduced, where target domain \textit{pseudo-}labels are estimated under an iterative self-training framework with spatial priors, achieving state-of-the-art results on the S2R adaptation task on SYNTHIA and GTA to CS experiments.

An orthogonal task to UDA is style transfer \cite{Gatys}, as the transfer of the style of a target domain to a source one, aligns with the concept of adapting a model from one domain to another.
This was the driving idea behind \cite{atapour2018real} which used a GAN to learn the task of monocular depth estimation in the target domain and showcased their concept in a S2R adaptation task using GTA and KITTI \cite{menze2015object}.
Nevertheless, transferring the style of one domain to another has only been an implicit goal of some approaches while it has been explicitly used as data augmentation and partly for domain adaptation by \cite{ArtisticVia, ArtisticBy}. 
Both utilize artistic -- and not photo-realistic -- style transfer, for improving the network's generalization. 
The former \cite{ArtisticVia} investigate the effect of style augmentation on three tasks (image classification, cross-domain classification and depth estimation) whereas the latter \cite{ArtisticBy} enforces the network to learn domain invariant features between source and restyled samples.
%The most related works to ours are \cite{SurReal} and the parallel archival works of \cite{jackson2018style} and \cite{dundar2018domain}. 
%The first one utilizes universal style transfer \cite{li2017universal} under label constraints to infuse realism into surgical simulation.
%The two latter ones explicitly use a style transfer method for infusing the style of the target domain to the source images and directly train a network to perform the task at hand using the restyled data.
%In \cite{jackson2018style} an artistic style transfer scheme is used with \cite{GoogleStyleTransfer} to augmented the dataset and increase performance for an image classification task, while also increasing robustness to domain shifts. 
%Contrary, a recently introduced fast style transfer method \cite{fastphoto} is used in \cite{dundar2018domain}.   
%However its focus is only on semantic segmentation with an iterative scheme for transferring the style relying on the labels.
%Nonetheless, results are offered for outdoors as well as indoors scenarios as well as object detection, using GTA, SYNTHIA, Cityscapes, KITTI, SunCG \cite{song2016ssc} and NYU \cite{NYU}.

Overall, we observe a trend in using adversarial learning and cycle consistencies for UDA, as well as a focus on classification, either image-wide or at the pixel level. 
Further, even though these methods' complexity is increasing, we rarely see evaluations on a wider domain spectrum, i.e. what is typically considered as ``\textit{real}" is a single dataset acquired from the real world, but the fact that there are still different ``\textit{real}" domains has not been properly addressed.
Some works have lightly touched this by evaluating their performance in unseen ``similar" domains \cite{atapour2018real, romijnders2018domain}, but most works typically require re-training to properly adapt to another ``\textit{real}" domain \cite{FCN}.

\section{Approach}
\label{sec:approach}

\newcommand{\Xa}{\pazocal{X}}
\newcommand{\Ya}{\pazocal{Y}}
\newcommand{\Za}{\pazocal{Z}}
\newcommand{\Ha}{\pazocal{H}}
\newcommand{\Ta}{\pazocal{T}}
\newcommand{\dH}{\pazocal{D_{H}}}

%In order to formulate the general problem of unsupervised domain adaptation we consider a set of input image source samples $x_s$ drawn from a distribution $\Xa_s$ along with their corresponding labelled data $y_s$.

The general problem of UDA considers a set of samples $x_s$ drawn from a \textit{source} distribution $\Xa_s$ along with their corresponding labels $y_s$.
%The latter can be single labels (image classification), image domain labels (semantic segmentation) or image domain ground truth values (depth estimation) or any other task dependant output.
For vision related tasks, the labels $y_s$ are either on a per image basis (e.g. classes or bounding boxes) or on a per pixel basis (e.g. class labels for semantic segmentation, continuous scalar or vector values for depth or surface estimation respectively) depending on the task.
%We also consider a set of input image target samples $x_t$ drawn from a distribution $\Xa_t$, with different characteristics than $\Xa_s$.
In addition, a \textit{target} distribution $\Xa_t$, different from $\Xa_s$, but similar in context, is also considered, offering a variety of unlabelled samples $x_t$.
%Our goal is to train a function $f$ with parameters $\theta$ in order to maximize the performance of $y_t = f(x_t, \theta)$ without the availability of corresponding labelled data $y_t$ for the target domain $\Xa_t$.
The goal is to train a machine learning model (CNN in the case of vision tasks) with parameters $\theta$, in order to learn a function $f$, without the availability of corresponding labelled data $y_t$ for the target domain $\Xa_t$, and maximize its performance when applied to data derived from $\Xa_t$.
%the performance of $y_t = f(x_t, \theta)$
Most approaches train a function $y = f(x, \theta)$ using the paired source samples ${x_s, y_s}$ and then adapt the parameters $\theta$ to increase performance on the target domain.

%A similar general problem is that of style transfer where an input content image $x_c$ needs to be transformed using the style of another image $x_{st}$ to produce a restyled image $x_{re}$ that preserves the content representation of $x_c$ and the style of $x_{st}$.
As aforementioned, $\Xa_s$ and $\Xa_t$ are similar in context, thereby depicting similar content and differing only in color distribution and patterns, as well as potentially in the noise levels (this is especially true for the realistic vs synthetic domains). 
Driven by this, we identify a new solution to the UDA problem, based on the orthogonal style transfer concept. 
Style transfer focuses on transferring the \textit{style} of one image $x_{st}$ to the \textit{content} of another image $x_c$, thereby producing a restyled image $x_{re}$ that preserves the content representation of $x_c$ and the style of $x_{st}$.
%This concept has been explored in previous works \cite{atapour2018real} but only as an implicit objective and not applied explicitly.

Our proposition stems from the intuition and observation that recent advances in style transfer better preserve the structure of the content image, while simultaneously transferring the appearance details of the style image. 
%We hypothesize that when training $f$ using the set of input restyled images $x_{re}$, the performance of $y_t = f(x_t, \theta)$ will increase.
We hypothesize that by enriching the source data distribution $\Xa_s$ with data drawn from a restyled distribution $\Xa_{re}$, and training $f$ with the new data distribution $\Za := \Xa_s \cup \Xa_{re}$, the performance of $y_t = f(x_t, \theta)$ will increase. 
%\textbf{According to \cite{Augmentation19}, it was proved that augmentations lead to feature averaging, improving generalization by inducing model invariance. Hence, we expect our nonlinear style transformations, which infuse the target domain’s features into the data, will improve the performance of our model.}
The restyled distribution will produce samples $x_{re}$ that will be the result of style transfer from a set of target (style) samples $x_t$ to a set of source (content) samples $x_s$.
As shown in \cite{Augmentation19}, data augmentations lead to feature averaging and improve generalization by inducing model invariance.
Hence, we expect that our non-linear style transfer transformations will infuse the target domain's features into the data and improve the performance of the model in the target distribution.
Consequently, $\Za$ will still depict the same content but it will also include the style of the target domain, enabling the learning of features aligned to both domains. 
An illustration of this concept can be found in Figure ~\ref{fig:system_overview}.
%mention -> Augmentations (stanford)

%Further, \cite{Augmentation19} recently showed that augmentations lead to feature averaging, improving generalization by inducing model invariance. This can explain how our non-linear style transformations, which infuse the target domain’s features into the data, offer the demonstrated performance gains.}
%These restyled input samples $x_{re}$ are produced by using style transfer on source images $x_s$ with a set of input style images $x_{st}$ drawn from a distribution $\Xa_{st}$ that can either be the target distribution $\Xa_t$ itself, or any other distribution $\Xa_o$ that is closer to the target distribution $\Xa_t$ than the source $\Xa_s$ one. 

%Under this hypothesis, the next natural step is the selection of style images. 
%However, style transfer operates on the basis of pre-defined content and style images.
%In the case of unsupervised domain adaptation, the selection of the samples that will be used to restyle the source data is an %open problem as correspondences cannot be easily established. 
%What is the most appropriate way to match content samples from the source distribution $\Xa_s$ with styles from the target %distribution $\Xa_t$?
A naive approach would be to randomly sample (RS) the target distribution $\Xa_{t}$ to find matching styles to each source sample.
However, that will lead to uncanny restyles and reduce the adaptation efficiency. 
%This can be alleviated by increasing the number of styles picked for each sample and thus, potentially reducing the percentage of uncanny restyles. 
%Driven by the same assumption that most domain adaption methods operate on - the fact that both the source and target domains depict similar content and in partly equal pixel and semantic distributions - we can base our selection process on this.
We propose to use the samples' structure to enforce consistency in the matching process and facilitate more effective data restyling to improve domain alignment.
%Naturally, in the image domain, low frequencies correspond to slowly varying information (smooth regions), while high frequencies to quickly varying information (edges).
%Therefore, the former comprise the image content (i.e. structure), while the latter correspond to the image's style (i.e. texture).
Under a spatial frequency transform, the image domain can be disentangled into the main structural information and its detail layers. 
Naturally, low frequencies correspond to the former and comprise the image's content, while higher frequencies composite texture onto the structure and comprise the image's style.
Therefore, after encoding all images (source and target alike) under a frequency transform and matching them using this encoding, the corresponding spatial structure information between the matches will produce visually appealing restyled results.

Accordingly, we leverage perceptual hashing (PH) \cite{phashing}, a technique used in reverse image search products (e.g. Google Images), to match source and target samples.
We transform samples from $\Xa_{s}$ and $\Xa_{t}$ using discrete cosine transform (DCT) into the frequency domain: $$\Ta = A \times g \times A^{T},$$
%\vspace{-3mm}
%\begin{equation}
    %\Ta = A \times g \times A^{T},
%\end{equation}
%\vspace{-3mm}
where $\Ta$ represents the resulting $DCT$ spectrum, A is a quadratic, real-valued, orthonormal transformation matrix, and $g$ is a low-resolution grayscale transformed sample $x$.
Then, we extract the low frequencies ($8\times8$ top left block) and calculate the median value after excluding the high amplitude DC term.
Subsequently, a $64$-bit hash $\Ha$ is constructed after thresholding the lowest frequencies based on the computed median value.
%Following the hashes calculation for all samples, 
Finally, structural matches can be established between the samples drawn from the source $\Xa_s$ and target $\Xa_t$ distributions, using their hashes' Hamming distance as a matching criterion.
It should be noted that unlike cryptographic hash functions, in perceptual hashing, small changes in the input will not lead to drastic changes in the output.
A qualitative comparison between randomly restyling data and selecting structural matches can be seen in Figure~\ref{fig:goodbad}, where it is apparent that matched restyles produce more plausible results when using Mapillary as the style domain.

\captionsetup[table]{skip=0pt} 
\begin{table*}[!t]
\centering
\centering
\caption{Domain adaptation results on the Mapillary Vistas dataset, As with Table \ref{table:gta2city}, the same applies regarding metrics and results presentation, except from ours, where the second row shows results for the configuration \srcTarget{GTA}{RS-Map}{Map}, and the third for \srcTarget{GTA}{PH-Map}{Map}.} \label{table:gta2map}

\srcTarget{GTA}{Mapillary}{Mapillary}
\resizebox{\linewidth}{!}{%
\begin{tabular}{c|c|ccccccccccccccccccc|cccc} \toprule
%\hline
%\\ %\midrule
    {$Method$} & {$network$} &
\begin{turn}{45}{$road$} \end{turn} & \begin{turn}{45}{$sidewalk$} \end{turn}&\begin{turn}{45} {$building$} \end{turn}&\begin{turn}{45} {$wall$} \end{turn}&\begin{turn}{45}{$fence$} \end{turn}&\begin{turn}{45} {$pole$} \end{turn} & \begin{turn}{45}{t-light} \end{turn} & \begin{turn}{45} {$t-sign$} \end{turn} & \begin{turn}{45}{$veg$} \end{turn} &\begin{turn}{45}{$terrain$} \end{turn}&\begin{turn}{45}{$sky$} \end{turn}&\begin{turn}{45}{$person$} \end{turn}&\begin{turn}{45}{$rider$} \end{turn}&\begin{turn}{45}{$car$} \end{turn}&\begin{turn}{45}{$truck$} \end{turn}&\begin{turn}{45}{$bus$} \end{turn}&\begin{turn}{45}{$train$} \end{turn}&\begin{turn}{45}{$mbike$} \end{turn}&\begin{turn}{45}{$bike$} \end{turn}& {$mIoU$}& {$pix.acc$} &{$mIoU gain$}&{$pix.acc.gain$}   \\ 
\hline

    \multirow{2}{*}{UADA \cite{romijnders2018domain}} & ResNet-50 &  &  &  &  &  &  & &  &  &  &  &  &  &  & & & &  & & 37.1&-& \\
	      &  &  &  &  &  &  &  & & &  &  &  &  &  &  & & & &  & & \textbf{38.5} & - &+1.4 & \\
	\hline 
     Baseline  &  & 48.83 & 19.24 & 59.37 & 6.47 & 12.82 & 13.55 & 14.62 & 13.77 & 64.63  & 10.51 & 89.63 & 35.8 & 13.25 & 43.7 & 21.32 & 7.81 & 1.0 & 22.66 & 18.11 & 27.22 & 72.4 \\
    Ours-RS& ResNet-101  & 72.88 & 24.68 & 66.74 & \textbf{14.93} & 17.13 & 15.14 & 19.83 & 26.12 & 72.49 & 16.44 & \textbf{92.74} & 37.63 & 9.07 & 71.77 & 33.6 & \textbf{12.93} & 0.0 & \textbf{25.56} & 19.39 & 34.16 & 81.1 & +6.94 &+8.7  \\
	     Ours-PH& & \textbf{82.71} & \textbf{28.01} & \textbf{68.81} & 13.88 & \textbf{18.87} & \textbf{18.18}  & \textbf{20.29} & \textbf{38.74} & \textbf{76.37} & \textbf{22.68} & {92.14} & \textbf{40.54} & \textbf{17.4} & \textbf{76.58} & \textbf{39.8} & 7.86 & \textbf{1.02} & 24.54 & \textbf{24.63} & 37.53 & \textbf{85.9} &\textbf{+10.31} &\textbf{+13.5} \\
 \hline
\end{tabular}}
%\vspace{0.1cm}
%\label{table:gta2map}
\end{table*}

\section{Experiments}
\label{sec:experiments}

%\input{tables/gta2Mapillary_IOU.tex}
%\small{
\begin{table*}[!ht]
\vspace*{0.5 cm}
 \centering
   \caption{Domain adaptation results on the Apolloscape dataset, showing IoU for each class, mean IoU (mIoU), pixel accuracy, mIoU gain and pixel accuracy gain.}
    \label{table:gta2apollo}
    \srcTarget{GTA}{Mapillary}{Apolloscape}
%\footnotesize{
    \resizebox{\linewidth}{!}{%
        \begin{tabular}{c|c|ccccccccccccccccccc|cccc}
        \hline
       \begin{turn}{0}{$Method$}\end{turn}&\begin{turn}{0}{$network$}\end{turn}&\begin{turn}{45}{$road$}\end{turn}& \begin{turn}{45} {$sidewalk$} \end{turn}&\begin{turn}{45} {$building$} \end{turn} & \begin{turn}{45} {$wall$} \end{turn} & \begin{turn}{45}{$fence$}\end{turn} & \begin{turn}{45} {$pole$}\end{turn} & \begin{turn}{45}{t-light}\end{turn} & \begin{turn}{45} {$t-sign$}\end{turn} & \begin{turn}{45}{$veg$}\end{turn} &\begin{turn}{45}{$terrain$}\end{turn}&\begin{turn}{45}{$sky$} \end{turn}&\begin{turn}{45}{$person$} \end{turn}&\begin{turn}{45}{$rider$} \end{turn}&\begin{turn}{45}{$car$} \end{turn}&\begin{turn}{45}{$truck$} \end{turn}&\begin{turn}{45}{$bus$} \end{turn}&\begin{turn}{45}{$train$} \end{turn}&\begin{turn}{45}{$mbike$} \end{turn}&\begin{turn}{45}{$bike$}\end{turn}& \begin{turn}{0}{$mIoU$}\end{turn}&\begin{turn}{0}{$pix.acc$}\end{turn}&\begin{turn}{0}{$mIoU gain$}\end{turn}&\begin{turn}{0}{$pix.acc.gain$}\end{turn} \\ 
        \hline
         \multirow{2}{*}{UADA \cite{romijnders2018domain}} & \multirow{2}{*}{\begin{turn}{0}{ResNet-50}\end{turn}} & & & & & & & & & & & & & & & & & & & &  25.3 & -  \\
         &  &  &  &  &  &  & & &  &  &  &  &  &  & & & &  & & & 27.4 & - & +2.1\\
	    \hline
      Baseline& & 80.52 & 0.19 & 7.16 & 0.14 & 18.12 & 7.16 & 6.22 & 2.82 & 41.89  & -  & 80.27 & 0.45  & \textbf{0.05} & 75.63 & 33.74 & 6.42 & - & 0.0 &  0.0 & 21.39 & 64.9 &  \\
     Ours-RS & \multirow{3}{*}[1em]{\begin{turn}{0}{{ResNet-101}}\end{turn}} &  83.52 & \textbf{0.22} & \textbf{15.38} & \textbf{0.48} & \textbf{19.17} & \textbf{14.06} & \textbf{16.12} & 7.53 & 72.54 & - & 89.46 & 0.43 & {0.01} & 73.59 & 45.61 & 7.31 & - & 0.0 & 0.0 & 26.2 &  81.3 & +4.81 & +16.4\\
 Ours-PH && \textbf{85.21}  & 0.17 & 10.85 & 0.42  & \textbf{19.17} & 11.2 & 14.46 & \textbf{40.56} & \textbf{74.63} & - & \textbf{90.31} & \textbf{0.9} & 0.0 & \textbf{76.86} & \textbf{62.54} & \textbf{30.27} & - & 0.0 & 0.0 & \textbf{30.44} & \textbf{82.9} & \textbf{+9.05} & \textbf{+18} \\
%VGG-16
%\hline
%      Baseline& & 69.6 & 0.5 & 14.7 & 1.8 & 33.4 & 16.0 & 5.3 & 15.6 & 77.7  & -  & 82.4 & 0.1  & 0.0 & 66.5 & 25.6 & 9.0 & - & 0.0 &  0.0 & 22.4 & 73.5 &  \\
%     Ours-RS & \multirow{3}{*}[1em]{\begin{turn}{0}{{VGG-16}}\end{turn}} &  73.5 & 0.7 & 17.0 & 1.5 & 33.0 & 18.3 & 4.6 & 9.0 & \textbf{78.3} & - & \textbf{84.6} & 0.1 & 0.0 & 67.1 & \textbf{34.2} & \textbf{12.3} & - & 0.0 & 0.0 & 25.5 & 82.0   & +3.0 & +8.6 \\
% Ours-PH && \textbf{86.5}  & \textbf{1.3} & \textbf{18.6} & \textbf{2.2}  & \textbf{33.9} & \textbf{18.6} & \textbf{5.7} & \textbf{22.3} & 77.4 & - & 83.0 & \textbf{0.4} & 0.0 & \textbf{67.2} & 33.7 & 10.8 & - & 0.0 & 0.0 & 27.2 & \textbf{82.8} & \textbf{+4.7} & \textbf{+10.4}\\

\hline
%\vspace{-3mm}
\end{tabular}
}
%\linebreak
%\linebreak
%\label{gta2apollo}
%}
\end{table*}
\setlength{\tabcolsep}{6pt}
\normalsize
%\vspace{2.5cm}
\captionsetup[table]{skip=0pt} 
\begin{table*}[!ht]
\vspace*{0.5 cm}
    \centering
       \centering
    \caption{Domain adaptation results on CS, showing IoU for each class, mean IoU (mIoU), pixel accuracy, mIoU gain and pixel accuracy gain. For each method the first row presents results when trained on the source dataset only, and tested on the target. The second presents results when restyled from the target domain, except from ours, for which the second row shows the configuration \srcTarget{GTA}{RS-CS}{CS}, and the third the \srcTarget{GTA}{PH-CS}{CS}.}
    \srcTarget{GTA}{Cityscapes}{Cityscapes}
    \resizebox{\linewidth}{!}{%
        \begin{tabular}{c | c | c c c c c c c c c c c c c c c c c c c | c c c c }
            \hline
            {$Method$} & {$network$} & \begin{turn}{45}{$road$} \end{turn} & \begin{turn}{45}{$sidewalk$} \end{turn}&\begin{turn}{45} {$building$} \end{turn}&\begin{turn}{45} {$wall$} \end{turn}&\begin{turn}{45}{$fence$} \end{turn}&\begin{turn}{45} {$pole$} \end{turn} & \begin{turn}{45}{t-light} \end{turn} & \begin{turn}{45} {$t-sign$} \end{turn} & \begin{turn}{45}{$veg$} \end{turn} &\begin{turn}{45}{$terrain$} \end{turn}&\begin{turn}{45}{$sky$} \end{turn}&\begin{turn}{45}{$person$} \end{turn}&\begin{turn}{45}{$rider$} \end{turn}&\begin{turn}{45}{$car$} \end{turn}&\begin{turn}{45}{$truck$} \end{turn}&\begin{turn}{45}{$bus$} \end{turn}&\begin{turn}{45}{$train$} \end{turn}&\begin{turn}{45}{$mbike$} \end{turn}&\begin{turn}{45}{$bike$} \end{turn}& {$mIoU$}& {$pix.acc$} &{$mIoU gain$}&{$pix.acc.gain$}  \\ 
            \hline
    \multirow{2}{*}{FCN in the wild \cite{inthewild}} & VGG-16 & 31.9 & 18.9 & 47.7 & 7.4  & 3.1 & 16.0 & 10.4 & 1.0 & 75.5 &13.0  & 58.9 & 36.0 & 1.0 & 67.1 & 9.5 & 3.7 & 0.0 & 0.0 & 0.0 & 21.1 & - &   &  \\                  
    & & 70.4 &32.4 & 62.1 & 14.9 & 5.4 & 10.9 & 14.2 & 2.7 & 79.2 & 21.3 & 64.6 & 44.1 & 4.2 & 70.4 & 8.0 & 7.3 & 0.0 & 3.5 & 0.0 & 27.1 & - & +6 &  \\
    \hline
    \multirow{2}{*}{Curriculum \cite{zhang2017curriculum}}  & VGG-16 & 18.1 & 6.8  & 64.1 & 7.3 & 8.7 & 21.0 & 14.9 &16.8 & 45.9 & 2.4  & 64.4 & 41.6 & 17.5 & 55.3 & 8.4  & 5.0  &\textbf{6.9} & 4.3  & 13.8 & 22.3 & - &  \\
     &  & 74.9 & 22.0 & 71.7 & 6.0 & 11.9 & 8.4 & 16.3 & 11.1 & 75.7 & 13.3 & 66.5 & 38.0 & 9.3  & 55.2 & 18.8 & 18.9 & 0.0 &\textbf{16.8} & 14.6 & 28.9 & - & +6.6 \\
    \hline
    
   \multirow{2}{*}{\cite{sankaranarayanan2018learning}}  & VGG-16 &73.5&  21.3&  72.3&  18.9&  14.3&  12.5&  15.1&  5.3&  77.2&  17.4&  64.3&  43.7&  12.8&  75.4& 24.8&  7.8&  0.0&  4.9&  1.8&  29.6 & - \\ 
    &  & \textbf{88.0} &  30.5 & 78.6 &25.2 & \textbf{23.5} & 16.7 &\textbf{23.5} & 11.6 & 78.7 &27.2 &71.9 &\textbf{51.3} &\textbf{19.5} &\textbf{80.4} & 19.8 & 18.3 & 0.9 & 20.8 &\textbf{18.4} &\textbf{37.1}&-& +7.5 \\
    \hline

          \multirow{2}{*}{CyCADA \cite{CyCADA}}   & VGG-16 & 26.0 & 14.9 & 65.1 & 5.5  & 12.9  & 8.9 & 6.0 & 2.5  & 70.0 & 2.9 & 47.0 & 24.5 & 0.0  & 40.0 & 12.1  & 1.5 & 0.0  & 0.0  & 0.0  & 17.9 & 54.0 &       &  \\ 
     &  & 85.2 & \textbf{37.2} & 76.5 & 21.8 & 15.0 & \textbf{23.8} & 22.9 & \textbf{21.5} & 80.5 & \textbf{31.3} & 60.7 & 50.5 & 9.0 & 76.9 & 17.1 & \textbf{28.2} & 4.5  & 9.8 & 0.4 & 35.4 & 83.6 & \textbf{+17.5} & +\textbf{29.6} \\
\hline
 Baseline&                                  & 73.6 & 18.4 & 70.0 & 9.0 & 11.4 & 17.7 & 7.0 & 5.2 & 70.8  & 7.4 & 63.6 & 39.8 & 1.9 & 76.2 & 14.7 & 3.0 & 0.0 & 2.1 & 0.0 & 25.9 & 75.0 \\
                                 %Ours-RS& VGG-16  & 86.5 &  29.7 &\textbf{79.4} & 21.75 & 13.82 &\textbf{26.81} & 7.79 & 2.48 &\textbf{79.51} & 20.51  & 70.4 & 45.7 & 3.98 & 78.58 &20.72 &\textbf{24.42} & 4.74 & 5.6 & 0.0 & 32.76  &83.96 & +6.84 &+8.94\\
	                              %Ours-PH&   & 86.12 & 30.5 & 77.57 & 19.3 & 9.56 & 20.69 & 7.37 & 9.07 & 79.26 & 23.2 & 71.52 & 46.13 & 3.01 & 80.11 & 22.78 & 13.5 & 0.09 & 5.48 & 0.0 & 32.06 &\textbf{84.12} & +6.14 & +\textbf{9.1}\\
Ours-RS& VGG-16  & \textbf{88.0} &  32.6 &80.0 & \textbf{28.6} & 20.2 &18.7 & 18.2 & 8.2 &\textbf{81.0} & 28.8  & 77.8 & 46.9 & 7.9 & 79.3 &\textbf{25.4} &27.3 & 1.7 & 9.6 & 0.0 & 35.8  &\textbf{85.9} & +9.9 &+10.9\\
Ours-PH&   & 86.3 & 31.4 & \textbf{80.3} & 28.2 & 11.7 & 22.9 & 20. & 12.8 & \textbf{81.0} & 26.7 & \textbf{78.4} & 47.9 & 11.4 & 79.1 & 22.7 & 17.0 & 4.0 & 11.4 & 0.0 & 35.5 &85.2 & +9.6 & +10.2\\
      
    \hline
     
     %\multirow{2}{*}{\cite{dropoutRegularization}}   & ResNet-50 \cite{resnet101} & 64.5 &24.9 &73.7& 14.8& 2.5& 18.0& 15.9& 0 &74.9& 16.4& 72.0& 42.3& 0.0& 39.5& 8.6& 13.4& 0.0 &0.0 &0.0 &25.3& -  \\ 
     %&  & 87.8& 15.6& 77.4& 20.6& 9.7& 19.0& 19.9& 7.7& 82.0& 31.5& 74.3& 43.5& 9.0& 77.8& 17.5& 27.7& 1.8& 9.7& 0.0& 33.3& - &+8 \\
    %\hline
    \hline
    \multirow{2}{*}{\cite{DAclassBalanced}}   & ResNet-38 \cite{resnet38} &70.0& 23.7& 67.8& 15.4& 18.1& 40.2& 41.9& 25.3& 78.8& 11.7& 31.4& \textbf{62.9}& \textbf{29.8}& 60.1& 21.5& 26.8& 7.7& \textbf{28.1}& 12.0& 35.4& -   \\ 
    &  &89.6 &\textbf{58.9}& 78.5& \textbf{33.0}& \textbf{22.3}& \textbf{41.4}& \textbf{48.2}& \textbf{39.2}& \textbf{83.6}& 24.3& 65.4& 49.3& 20.2& \textbf{83.3}& \textbf{39.0} &\textbf{48.6} &\textbf{12.5}& 20.3& 35.3& \textbf{47.0} & - & +11.6 & \\
    \hline
    \multirow{2}{*}{I2I Adapt \cite{im2im2}}           & {DenseNet} \cite{densenet} & 67.3 & 23.1& 69.4 &13.9  &14.4  &21.6  &19.2 &12.4 &78.7 &24.5  &74.8  &49.3  &3.7  &54.1  &8.7 &5.3 & 2.6&6.2  &1.9 & 29.0 & -\\
	               & & 85.8  &37.5  &80.2  &23.3  & 16.1 & 23.0 & 14.5 & 9.8 &79.2  &\textbf{36.5}  &76.4 &53.4  &7.4  &82.8 &19.1 &15.7 &2.8  &13.4 &1.7 &35.7 & - & +6.7 \\
    \hline
    \multirow{2}{*}{SG-GAN \cite{semAwareGradGAN}}        &  PatchGAN\cite{patchGAN} & & & & & & &  & & & & & & & & & & & & & 24.6 & 54.5 \\
                         & & &  &  & & & &  &  & & & & & & & & & & & & 37.4 & 81.72 & +\textbf{12.8} & \textbf{+27.2}\\
    \hline
    \multirow{2}{*}{UADA \cite{romijnders2018domain}}                      & ResNet-50 &  &  &  &  &  &  & &  &  &  &  &  &  & & & &  & & & 34.8 & - \\
	                          &  &  &  &  &  &  &  & &  &  &  &  &  &  & & & &  & & & 38.2 & - & +3.4 \\
	\hline
% Baseline&                                  & 64.27 & 17.98 & 75.83 & 12.16 & 12.35 & 14.02 & 7.78 & 6.65 & 75.73  & 9.73 & 75.56 & 49.36 & 14.09 & 39.08 & 9.03 & 16.11 & 0.0 & 12.78 & 18.77 & 27.96 & 68.2 \\
     Baseline&                                  & 47.9 & 19.3 & 66.4 & 11.8 & 10.9 & 23.8 & 12.4 & 15.5 & 79.0  & 17.3 & 75.3 & 53.5 & 17.4 & 72.6 & 6.3 & 1.8 & 0.1 & 16.5 & 24.9 & 30.2 & 66.8 & &\\
                                 Ours-RS& ResNet-101  & \textbf{90.1} &  38.1 & \textbf{80.4} & 15.2& 14.5 & 28.4 & 13.9 & 19.9 & 81.7 & 31.4  & \textbf{78.4} & 51.6 & 18.1 & 78.0 & 20.3 & 22.5 & 1.4 & 18.4 & \textbf{38.4} & 39.0  &\textbf{86.3} &+8.8  &+19.5\\
	                              Ours-PH& &89.3  & 47.1 & 80.3 & 13.2 & 13.4 & 20.1 & 13.4 & 19.2 & 82.5 & 29.6 & 77.2 & 52.3 & 16.0 & 80.4 & 19.0 & 25.1 & 1.84 & 15.0 & 30.1 & 38.2 & 85.9 &+8.0 &+19.1\\
	\hline
\end{tabular}}

\label{table:gta2city}
\end{table*}
\begin{table*}[!ht]
\vspace*{0.5 cm}
    \centering
       \caption{Domain adaptation results on CS, showing IoU for each class, mean IoU (mIoU), pixel accuracy, mIoU gain and pixel accuracy gain. For each method the first row presents results when trained on the source dataset only, and tested on the target. The second presents results when restyled from the target domain, except from ours, for which the second row shows the configuration \srcTarget{GTA}{RS-Map}{CS}, and the third the \srcTarget{GTA}{PH-Map}{CS}}
    \srcTarget{GTA}{Mapillary}{Cityscapes}
    \resizebox{\linewidth}{!}{%
       \begin{tabular}{c | c | c c c c c c c c c c c c c c c c c c c | c c c c }
            \hline
            {$Method$} & {$network$} & \begin{turn}{45}{$road$} \end{turn} & \begin{turn}{45}{$sidewalk$} \end{turn}&\begin{turn}{45} {$building$} \end{turn}&\begin{turn}{45} {$wall$} \end{turn}&\begin{turn}{45}{$fence$} \end{turn}&\begin{turn}{45} {$pole$} \end{turn} & \begin{turn}{45}{t-light} \end{turn} & \begin{turn}{45} {$t-sign$} \end{turn} & \begin{turn}{45}{$veg$} \end{turn} &\begin{turn}{45}{$terrain$} \end{turn}&\begin{turn}{45}{$sky$} \end{turn}&\begin{turn}{45}{$person$} \end{turn}&\begin{turn}{45}{$rider$} \end{turn}&\begin{turn}{45}{$car$} \end{turn}&\begin{turn}{45}{$truck$} \end{turn}&\begin{turn}{45}{$bus$} \end{turn}&\begin{turn}{45}{$train$} \end{turn}&\begin{turn}{45}{$mbike$} \end{turn}&\begin{turn}{45}{$bike$} \end{turn}& {$mIoU$}& {$pix.acc$} &{$mIoU gain$}&{$pix.acc.gain$}  \\ 
  \hline
%\multirow{3}{*} \\ Baseline&                                  & 73.6 & 18.4 & 70.0 & 9.0 & 11.4 & 17.7 & 7.0 & 5.2 & 70.8  & 7.4 & 63.6 & 39.8 & 1.9 & 76.2 & 14.7 & 3.0 & 0.0 & 2.1 & 0.0 & 25.9 & 75.0 \\
                                 Ours-RS& VGG-16  & 86.5 &  29.7 &\textbf{79.4} & \textbf{21.7} & \textbf{13.8} &\textbf{26.8} & \textbf{7.7} & 2.4 &\textbf{79.5} & 20.5  & 70.4 & 45.7 & \textbf{3.9} & 78.5 &20.7 &\textbf{24.4} & \textbf{4.7} & \textbf{5.6} & 0.0 & \textbf{32.7}  &83.9 & \textbf{+6.8} &+8.9\\
	                              Ours-PH&   & \textbf{86.1} & \textbf{30.5} & 77.5 & 19.3 & 9.5 & 20.6 & 7.3 & \textbf{9.0} & 79.2 & \textbf{23.2} & \textbf{71.5} & \textbf{46.1} & 3.0 & \textbf{80.1} & \textbf{22.7} & 13.5  & 0.0& 5.4 & 0.0 & 32.1 &\textbf{84.1} & +6.1 & +\textbf{9.1}\\
            \hline
         %   \multirow{3}{*} \\ Baseline&                                  & 64.2 & 17.9 & 75.0 & 12.1 & 12.3 & 14.0 & 7.8 & 6.6 & 75.7  & 9.7 & 75.5 & 49.3 & 14.0 & 39.1 & 9.0 & 16.1 & 0.0 & 12.7 & 18.7 & 27.9 & 68.2 \\
                                 Ours-RS& ResNet-101  & 87.5 &  22.1 & \textbf{80.8} & \textbf{18.8} & \textbf{17.3} & 15.4 & \textbf{10.0} & 7.4 & 81.5 & 20.3  & 81.4 & \textbf{50.5} & \textbf{19.5} & \textbf{77.4} & 26.5 & 29.3 & 0.1 & 16.1 & 16.6 & 35.7  &\textbf{86.0} & +5.5 &\textbf{+19.2}\\
	                              Ours-PH&   & \textbf{87.7} & \textbf{24.6} & 79.0 & 16.0 & 7.9 & \textbf{16.7} & 7.0 & \textbf{8.9} & \textbf{81.6} & \textbf{27.2} & \textbf{82.3} & 49.5 & 13.9 & 77.3 & \textbf{31.3} & \textbf{38.3} & \textbf{0.5} & \textbf{25.7} & \textbf{18.5} & \textbf{36.5} & 85.6 & \textbf{+6.3} & + 18,8\\
	\hline

	%\hline
  %   \multirow{3}{*} \\ Baseline&                                  &27.96   &68.2  \\
   %                              Ours-RS& ResNet-101  & 35.04   & 84.49 & +7.08 & +16.45\\
	%                              Ours-PH&   & 35.69 & 85.6 & {+7.73} & {+17.4} \\
	%\hline
%VGG-16
%\vspace{-3mm}
\end{tabular}
}
\label{tab:exps}
\end{table*}

\begin{figure*}[!h]
    \centering
    \includegraphics[width=0.92\linewidth]{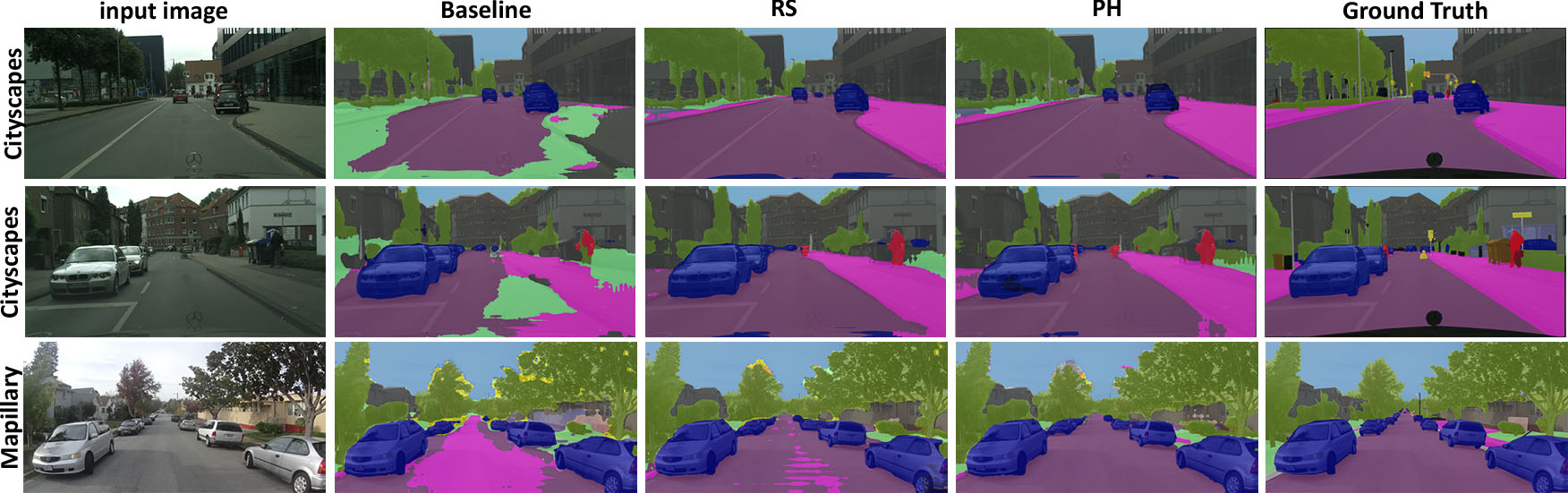}

    \caption{Qualitative results on adaptation from GTA5 to Mapillary and Cityscapes. Rows correspond to predictions for sample images in Cityscapes and Mapillary. From left to right: input image, baseline prediction, ours-RS and ours-PH. Best viewed in color.}
    \label{fig:fig1} % I can do without the label too
\end{figure*}

\begin{figure*}[!htbp]
    \centering
    \includegraphics[width=1.05\linewidth]{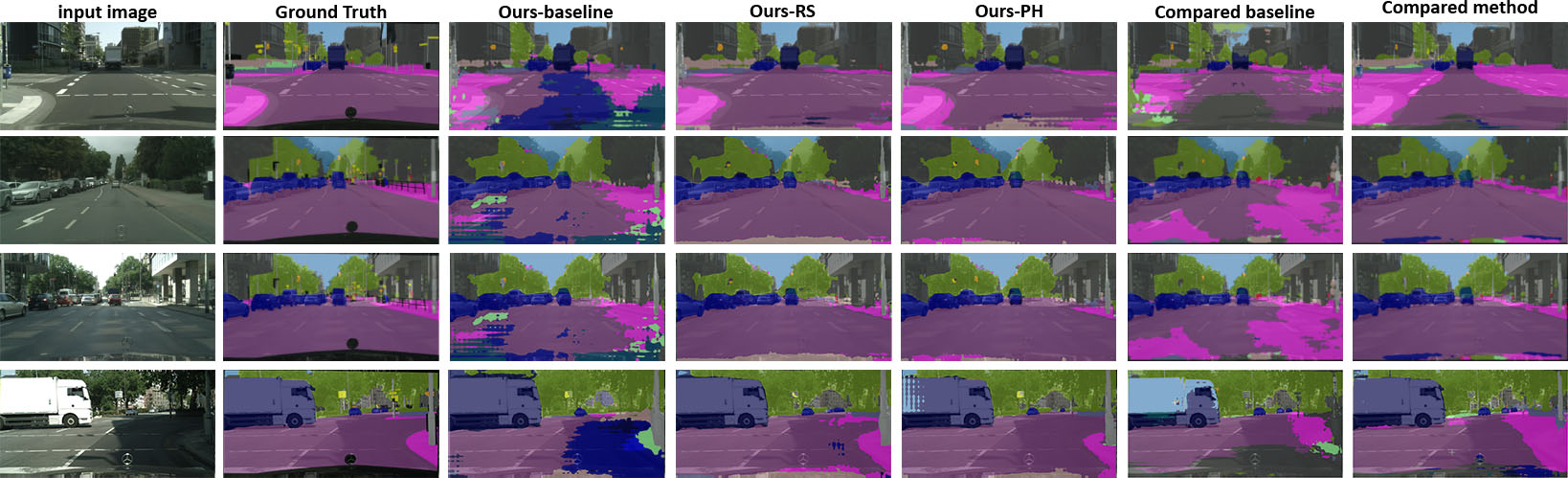}
    \caption{Qualitative results on samples of the Cityscapes dataset, for the semantic segmentation task on different datasets. From left to right: input rgb, our baseline result, our result with random style selection (RS), our result with style selection via perceptual hashing (PH), compared method's baseline result, and compared method's result. Each row presents comaprison with a different method. From top to bottom: \cite{DAclassBalanced}, SG-GAN \cite{semAwareGradGAN}, UADA \cite{romijnders2018domain}, and once more \cite{DAclassBalanced}.}
    \label{fig:segSemQualComp}
\end{figure*}

%In the following sections we seek to validate our hypothesis under a variety of different tasks $T$. 
Although our proposed technique is only applicable to image data, and thus to vision related tasks, it is conceptually generic. 
We conduct a series of experiments in two different vision tasks to validate this, mostly focusing on S2R UDA.
%We employ the state-of-the-art style transfer method of \cite{fastphoto} that better preserves structural information and is faster than other approaches.
%For image encoding we use a hashing method used in reverse image products such as Google Images \footnote{\href{https://images.google.com/}{https://images.google.com/}} and TinEye\footnote{\href{https://www.tineye.com/}{https://www.tineye.com/}}. 
%This perceptual hashing method \cite{phashing} is generated from multiple features extracted from a DCT 
%and are cryptography hashed 
%\textbf{so that unlike cryptographic hash functions, small changes in the input won't lead to drastic changes in the output. }
%This makes perceptual hashes employable for signatures used in image and video identification systems.

We define the following notation to disambiguate each dataset's role in the experiments: \srcTarget{\textit{src}}{\textit{stl}}{\textit{tgt}}, with \textit{src} referring to the source dataset, \textit{stl} to the dataset that the styles are derived from, and \textit{tgt} is the target dataset, i.e. the dataset used for our experimental validation.

\textbf{Implementation details:}
For style transfer we use a photo-realistic method \cite{fastphoto} that better preserves structural information and is faster than other approaches.
We apply data restyling as a pre-processing step on a source dataset and then merge the restyled data with the source ones, and use the merged dataset for training.
We first hash all source and target domain image samples and then use a brute-force nearest neighbor to match $K$ target samples to each source image.
%These are then used to produce the restyled source image dataset that is used for training the function $f_T$ for the task at hand.

Our experiments were run on a single machine with an i7 processor and a NVidia Titan X GPU.
Nonetheless, this technique is inherently parallel as all pre-processing steps (hashing, matching, restyling) can be run independently for each sample.
While the experiments on a single machine were lengthy, the proposed technique can easily scale out to allow for faster dataset-wide pre-processing or the accommodation of very large style/target datasets.
Setting out to validate and assess our proposed technique, we fix $K$ to $5$ for all conducted experiments, which is a good compromise between efficiency and effectiveness.
\subsection{Semantic Segmentation}
%\textbf{Semantic Segmentation:}
%Our next experiment is in line with a lot of the recent work done for unsupervised domain adaptation, as it focuses on outdoors semantic segmentation, and more specifically in driving scenarios. 
%After validating the concept, we present results for data restyling on the semantic segmentation task in a synthetic-to-real adaptation setting.
%Contrary to image classification that focuses on global image properties, semantic segmentation features need to encode much more complex relationships.
%We focus on virtual-to-real adaptation and 
%We expect that our technique will add important diversity to the virtual source domain and infuse it with properties from the real target domain. 
%We use four datasets, one virtual and three real:
%We use GTA as our labelled synthetic source data distribution, Mapillary, Apolloscape and Cityscapes for the real domain data distributions, and use the overlapping labels among them, as well as their official train/test splits.
%As aforementioned, we enrich the source dataset with our restyled samples and increase its size five-fold.

In this section, we present results for data
restyling on the semantic segmentation task in a synthetic to-real adaptation setting. We expect that our technique will add important diversity to the virtual source domain and infuse it with properties from the real target domain.

We use four datasets, one virtual and three real:

1) Cityscapes, a real-world dataset, capturing street scenes of $50$ different cities, totaling $5,000$ images with pixel-level labels. 
The dataset is divided into a training set with $2,975$ images, a validation set with $500$ images. 
All images are shot with the same car and camera.

2) GTA, a virtual dataset that contains $24,966$ high-resolution images, automatically annotated into $19$ classes. 
The dataset is rendered from the video-game Grand Theft Auto V and is based on the city of Los Angeles with labels fully compatible with those of Cityscapes.

3) Mapillary Vistas, a large-scale street-level image dataset containing $25,000$ high-resolution
images (split into $18,000$ for training, $2,000$ for validation, $5,000$ for testing; at an average resolution of ~9 megapixels). 
Images are shot in various countries using multiple cameras and their dense labels cover $66$ classes.

4) ApolloScape, an open source dataset collected at Beijing in China, including $25$ pre-defined semantic classes. 
Image frames in the dataset are collected every one meter with resolution $3384 \times 2710$. 
Among all available images, we have kept $8327$ frames for evaluation.

We use GTA as our labelled synthetic source data distribution, Mapillary, Apolloscape and Cityscapes for the real domain data distributions, and use the overlapping labels among them.
As aforementioned, we enrich the source dataset with our restyled samples and increase its size five-fold.

Our experiments are based on DeepLab-v2 \cite{deeplab} with ResNet-101 \cite{resnet101} and VGG-16 \cite{VGG} encoders, both initialized with pre-trained weights on ImageNet \cite{imagenet}.
All models are trained for $27$ epochs using SGD with a learning rate of $2.5 \times 10^{-4}$, supported by a $0.9$ momentum, a weight decay of $5 \times 10^{-4}$ and a polynomial learning rate decay policy with $\gamma=0.9$. 
Experiments are run using Tensorflow $1.10$ \cite{tensorflow} and PyTorch 1.1 \cite{pytorch}.
\vspace{0.2cm}
\begin{figure*}[!ht]
    \centering
    \begin{rotate}{90} \small{Input RGB} \end{rotate}
    \begin{subfigure}[t]{0.3\linewidth}
        \includegraphics[scale=1]{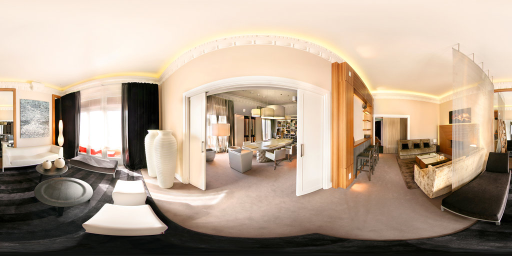}
    \end{subfigure}
    \begin{subfigure}[t]{0.3\linewidth}
        \includegraphics[scale=0.8]{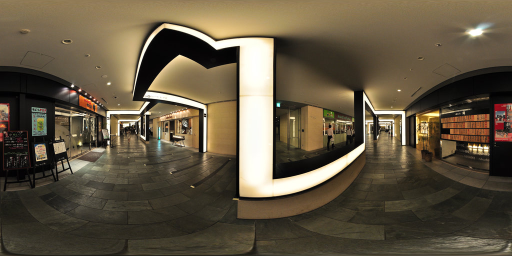}
    \end{subfigure}
    \begin{subfigure}[t]{0.3\linewidth}
        \includegraphics[scale=0.24]{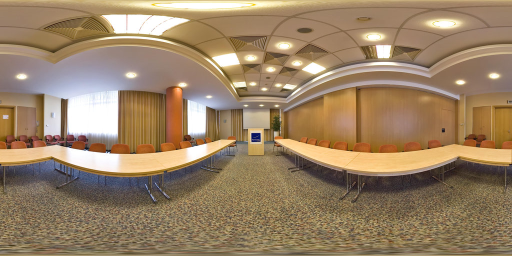}
    \end{subfigure}
    
    \begin{rotate}{90} \small{Source} \end{rotate}
    \begin{subfigure}[t]{0.3\linewidth}
        \includegraphics[scale=0.24]{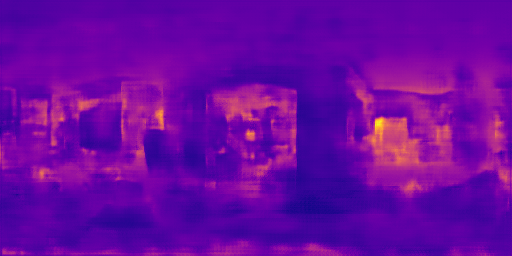}
    \end{subfigure}
    \begin{subfigure}[t]{0.3\linewidth}
        \includegraphics[scale=0.24]{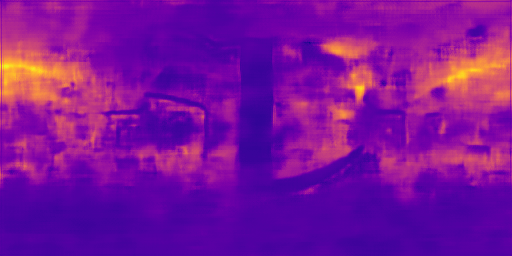}
    \end{subfigure}
    \begin{subfigure}[t]{0.3\linewidth}
        \includegraphics[scale=0.24]{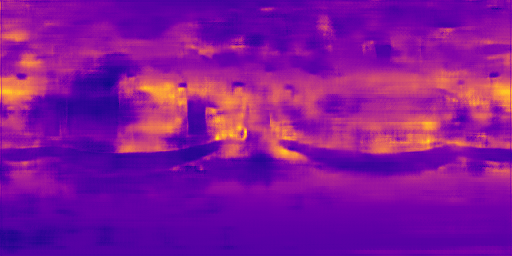}
    \end{subfigure}
    
%    \begin{rotate}{90} \tiny{RS} \end{rotate}
%    \begin{subfigure}[t]{0.3\columnwidth}
%        \includegraphics[scale=0.12]{figures/img/depthEstimation_res/rs_pano_aantwokdeshnqe.png}
%    \end{subfigure}
%    \begin{subfigure}[t]{0.3\columnwidth}
%        \includegraphics[scale=0.12]{figures/img/depthEstimation_res/rs_pano_aadajgqflfzxta.png}
%    \end{subfigure}
%    \begin{subfigure}[t]{0.3\columnwidth}
%        \includegraphics[scale=0.12]{figures/img/depthEstimation_res/rs_pano_aagyhxxyxetsbz.png}
%    \end{subfigure}
    
    \begin{rotate}{90} \small{PH} \end{rotate}
    \begin{subfigure}[t]{0.3\linewidth}
        \includegraphics[scale=0.24]{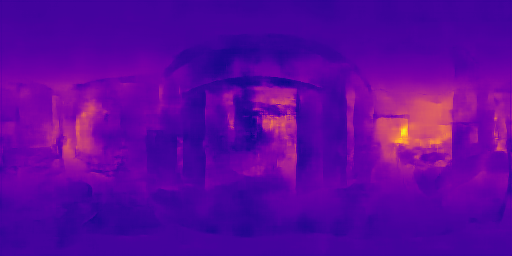}
    \end{subfigure}
    \begin{subfigure}[t]{0.3\linewidth}
        \includegraphics[scale=0.24]{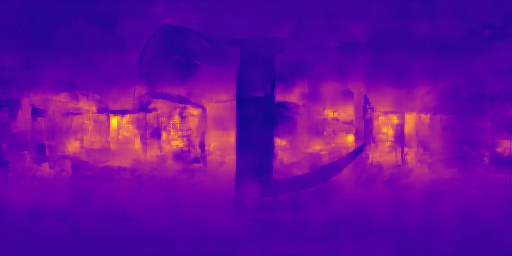}
    \end{subfigure}
    \begin{subfigure}[t]{0.3\linewidth}
        \includegraphics[scale=0.24]{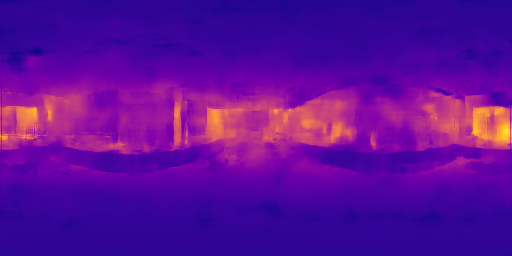}
    \end{subfigure}
    \caption{Qualitative results on omnidirectional dense depth estimation on samples of SUN360 where there is no ground truth depth map available.}
    \label{fig:depthEst}
\end{figure*}
\vspace{-3mm}

\vspace{0.9cm}

Table \ref{table:gta2map} presents results for \srcTarget{GTA}{Mapillary}{Mapillary} in comparison to \cite{romijnders2018domain}, the only method to presents results in Map, albeit in their case it is considered as unseen. 
``Unseen'' refers to the scenario where a domain adaptation model was trained using a specific target dataset to align its source domain to, but applied to another one, which is similar in context and represents the same domain as the target. It is a challenging scenario that has been lightly addressed in recent work.

We observe that data restyling offers competitive performance and a much larger increase compared to the baseline, with matched restyling (PH) additionally offering a higher boost compared to random selection (RS).
%While Map and CS are real domain datasets, their distributions are different, a fact that introduces performance deviations. 

Nonetheless, adapting data from the virtual domain to the real should not be bound to a specific dataset.
Whereas the use of unlabeled styles images from the target domain contributes efficiently to domain alignment, we should accord special respect to the case where the target distribution is partly different but of the same context. 
Owing to the fact that real-life scenarios consist of such cases, it is crucially important for domain adaptation methods and models to adapt properly to new, previously unseen data.
Hence, we take a step forward, and evaluate our technique in similar - but unseen - distributions as well.

%We conduct the \srcTarget{GTA}{Map}{AS} experiment and present its results in Table \ref{table:gta2apollo}.
Table \ref{table:gta2apollo} presents results for the conducted \srcTarget{GTA}{Map}{AS} experiment.
%We test our restyling technique on the AS validation test comprising $8327$ samples.
%Again, we use overlapping classes and test our restyling technique on the AS validation test comprising $8327$ samples.
While the synthetic data were restyled using Map's data distribution, the model's performance on the AS real -- but unseen -- data distribution is preserved and surpasses that reported in \cite{romijnders2018domain}.
As with \srcTarget{GTA}{Map}{Map}, a performance increase is observed for PH compared to RS.

Finally, we perform the \srcTarget{GTA}{CS}{CS} experiment and present our results in Table \ref{table:gta2city} which are consistent for both backends used. 
Since most recent works offer results on CS as well, we can compare our performance to the state-of-the-art. 
As some methods use different baselines, we report these and then focus on the gain offered by each method and find that our data restyling technique remains competitive with most similar methods.
Interestingly, our approach only pre-processes and enriches the data fed to the baseline model, while the compared methods utilize complex training schemes, embed indirect objectives and employ difficult to train GANs.
%\textcolor{red}{We also offer results for both backends and observe a discrepancy between them ...}

Curiously, unlike all other experiments, RS performance is higher than PH in CS, which can be attributed to CS's lack of diversity compared to the other datasets \cite{apollo}.
Table \ref{tab:exps} complements this final experiment by presenting results when using Map as the style pool.
We find that while performance degrades compared to using the target dataset (CS) as the style domain, there is still a gain when applied to another real data distribution.
Notably, the discrepancy between RS and PH is still different from all other experiments in that no significant increase is observed for PH when targeting CS. 
Apart from the quantitative evaluation, we also offer a set of exemplary qualitative results for both Map and CS in Figure \ref{fig:fig1}.
%Further, Table \ref{table:gta2city} presents results for the \srcTarget{GTA}{Mapillary}{CityScapes} unseen task. 
We additionally offer qualitative results on samples presented by \cite{DAclassBalanced, semAwareGradGAN, romijnders2018domain} in Figure \ref{fig:segSemQualComp}, showcasing the competitiveness of our technique.

Overall, we observe: \textbf{i)} that data restyling is competitive to most other compared UDA methods 
%\cite{romijnders2018domain, zhang2017curriculum, inthewild, im2m, semAwareGradGAN,dropoutRegularization, DAclassBalanced, sankaranarayanan2018learning, CyCADA} 
for semantic segmentation, \textbf{ii)} an increased generalization performance to unseen distributions, and \textbf{iii)} that matching styles helps in addressing the performance deterioration that uncanny restyles via random matching would introduce.

\subsection{Depth Regression}
%\textbf{Depth Regression:}
%Our final experiment considers an usual task for unsupervised domain adaptation, monocular dense depth regression from a single image which is an ill-posed problem that is also very difficult to acquire high quality labelled data as well. To that end, synthetic data generation is a very promising approach, making the virtual-to-real domain adaption a very relevant topic.  
%we show results for two domain adaptation scenarios, evaluated using the SUN360 dataset \cite{xiao2012recognizing}, the 360D dataset and the RectNet CNN architecture presented in.
\begin{table*}[!t]
\vspace*{1 cm}
    \centering
    \caption{Results of omnidirectional depth estimation for the different domain adaptation scenarios ($\downarrow$ means lower is better, and $\uparrow$ means higher is better). The first three rows represent evaluating on M3D and S2D3D datasets, while the latter two correspond to evaluations on the RS-M3D-S2D3D with styles derived form SUN360. Baseline implies that RectNet was trained on SunCG only, ours-RS that RectNet was trained on RS-SunCG, while ours-PH that RectNet was trained with PH-SunCG with styles from SUN360.
    \vspace*{1 cm}}
    %\tiny{
        \setlength{\tabcolsep}{3pt}
        \begin{tabular}{ l | c | c | c | c | c | c | c}
        \hline
        Method & 
        \makecell{Abs. Rel.$\downarrow$} & 
        \makecell{Sq. Rel.$\downarrow$} & 
        \makecell{RMSE $\downarrow$} & 
        \makecell{RMSE (log)$\downarrow$} & 
        \makecell{$\delta < 1.25$ $\uparrow$} & 
        \makecell{$\delta < 1.25^{2}$ $\uparrow$ } &
        \makecell{$\delta < 1.25^{3}$ $\uparrow$ }\\
        \hline
        \hline
        Baseline    & 0.233 & 0.204 & 0.669 & 0.283 & 0.614 & 0.891 & 0.966 \\
        Ours-RS     & 0.268 & 0.306  & 0.757 & 0.321 & 0.626 & 0.856 & 0.938 \\
        Ours-PH     & \textbf{0.209} & \textbf{0.188}  & \textbf{0.622} & \textbf{0.261} & \textbf{0.693} & \textbf{0.9} & \textbf{0.964} \\
        \hline
        Baseline    & 0.406 & 0.511 & 1.025 & 0.472 & 0.371 & 0.659 & 0.839 \\
        Ours-RS     & 0.264 & 0.287  & 0.747 & 0.318 & 0.626 & 0.857 & 0.936 \\
        Ours-PH     & \textbf{0.231} & \textbf{0.229}  & \textbf{0.681} & \textbf{0.285} & \textbf{0.666} & \textbf{0.884} & \textbf{0.956} \\
        \hline
    \end{tabular}
    %}   %   !tiny
    \label{tab:rectnet_da_results}
\end{table*}

We conduct a final experiment on a dense regression task, monocular depth estimation, an ill-posed problem that is furthermore very difficult to collect high quality labelled data.
Hence, using synthetic data generation is a very promising approach, constituting synthetic-to-real adaptation very relevant. 

We use the recently released 360D dataset \cite{Zioulis_2018_ECCV} that was generated via rendering synthetic, as well as realistic scanned scenes.
It comprises \360 renders of existing 3D datasets, annotated with their corresponding ground truth depth maps. 
It is composed of omnidirectional renders from Matterport3D (\textit{M3D}) \cite{Matterport3D}, Stanford2D3D (\textit{S2D3D}) \cite{2017arXiv170201105A} and SunCG \cite{song2016ssc} (\textit{SunCG}).
While the first two contain scanned 3D textured models of real indoors spaces, the third is composed of Computer Generated (synthetic) scenes only. 
In addition, the SUN360 \cite{xiao2012recognizing} dataset contains a very large variety of only color images, depicting real world scenes and captured with real \360 cameras. 
Given that the context of 360D is indoors scenes, we only use that subset of Sun360 for deriving realistic styles that consists of $33512$ samples.
%that are not paired with any ground truth annotation. 
%Likewise, we use a subset of the 360D dataset as there are a lot of invalid samples with missing information (e.g. renders with holes/missing regions due to incomplete CAD models, or from imperfect large scale scanning).
%This subset is formed by considering samples with $5\%$ or more of their depth map values exceeding $8m$ or being below $0.5 m$ as outliers. 
%We also use the valid samples of \cite{Zioulis_2018_ECCV} as explained

We use the synthetic SunCG dataset as our source domain and the realistic Sun360 dataset as our style pool.
We train RectNet \cite{Zioulis_2018_ECCV} only on the valid subset of SunCG ($4716$ samples) and consider it as our baseline.
%We pick and match $K=5$ Sun360 samples as styles for each SunCG source image after extracting their perceptual hashes.
We follow the same training and data validity schemes as in \cite{Zioulis_2018_ECCV}.
%For both datasets we train using the same network for 10 epochs using Caffe \cite{jia2014caffe} and the Adam solver \cite{kingma2014adam} for optimization, with its default parameters and a fixed learning rate of $0.0002$. 
%The training time of \srcTarget{SunCG}{}{M3D, S2D3D} is about 12 hours, while for the rest of the configurations is 70 hours.

%Despite using SUN360 as our target dataset, it does not contain any ground truth depth annotations. Because of this fact, we provide only qualitative results of our approach on SUN360. 
Given that Sun360 does not offer ground truth depth measurements we offer qualitative results of the baseline and domain adapted networks in Figure \ref{fig:depthEst}. It is evident that visually performance increases when applied to the unlabelled Sun360 domain directly.
In order to present quantitative results we also evaluate our trained models in an unseen task, i.e. \srcTarget{SunCG}{Sun360}{S2D3D-M3D} using the realistic S2D3D-M3D data.
%, which albeit from a different domain where performance degradation has been observed \cite{Zioulis_2018_ECCV}, are also "real" domain data.
In addition, we restyle the S2D3D-M3D data using the Sun360 style pool in order to infuse its style into the ground truth paired S2D3D-M3D data and be able to calculate quantifiable results.
We refer to this task as \srcTarget{SunCG}{Sun360}{RS-S2D3D-M3D}.
%The former contains $33512$ samples, while the latter contains a subset of $3917$ samples, which were randomly chosen, each restyled with $K=5$ random styles from Sun360, resulting to 19855 samples in total and is denoted it as \textbf{RS}-S2D3D-M3D.
%To sum up, we present results for the following configurations: \srcTarget{SunCG}{}{S2D3D-M3D}, \srcTarget{SunCG}{}{\textbf{RS}-S2D3D-M3D}, \srcTarget{SunCG}{\textbf{PH} - SUN360}{S2D3D-M3D} and \srcTarget{SunCG}{\textbf{PH}-SUN360}{\textbf{RS}-S2D3D-M3D} 

Quantitative results are presented in Table \ref{tab:rectnet_da_results}, for which we used typical metrics as reported in \cite{Zioulis_2018_ECCV}. 
The results further support previous findings in that restyling the data increases performance to both the targeted domain (\textbf{RS}-S2D3D-M3D), as well as similar domains (S2D3D-M3D), with the largest performance gained obtained on the domain that the styles were picked from.
We also find that the matched sample selection strategy helps in increasing domain adaptation performance compared to a random selection.

%-----------------------------------------------------------------------------------------------------
%\subsection{Generalization}
%\input{Generalization.tex}

%\section{Discussion}
%\label{sec:discussions}
%\textbf{Image resolution}
%Top performing approaches on CITYSCAPES, trained on synthetic data mostly use a high resolution for %better performance as reported in ... 
\section{Conclusion}
\label{sec:conclusion}
%\item See how this interplays with traditional GAN based domain adaptation 
We have shown that explicit data restyling can be an easy to implement, scalable and effective technique for UDA. 
Besides offering a straightforward way to avoid the complexity of adversarial based and cycle consistency based domain adaptation, it is also ripe for experimentation in conjunction with traditional methods, as it is purely a data pre-processing step.
However, it is because to this reason alone that it is much more applicable and, due to its inherently parallel nature, scales extremely well.
%In addition, it is fully parallelizable, and thus scalable. 
It poses as a very practical tool to adapt models when it is easy to obtain large quantities of unlabelled data (e.g. industrial/production settings).
%machine learning applications that can easily
%In contrast to recent work that makes extensive use of adversarial training, the demonstrated technique does not require focus on the domain adaptation task, allowing for purely engineering solutions for the task at hand. 
Given the generality of the proposed technique, it was validated in different vision tasks and datasets, and found to be a promising alternative to most approaches found in the literature, whilst not being task or dataset constrained.
% limitation -> per class IoU shows that we get better performance only on the "bigger" classes, future work to improve it on "smaller" classes as well
% Map distribution further than CS, AS closer, thus we observe good generalisation behavior

%-------------------------------------------------------------------------
{\small
\bibliographystyle{ieee}
\bibliography{egbib}

\begin{thebibliography}{10}\itemsep=-1pt

\bibitem{tensorflow}
M.~{Abadi}, A.~{Agarwal}, P.~{Barham}, E.~{Brevdo}, Z.~{Chen}, C.~{Citro},
  G.~S. {Corrado}, A.~{Davis}, J.~{Dean}, M.~{Devin}, S.~{Ghemawat},
  I.~{Goodfellow}, A.~{Harp}, G.~{Irving}, M.~{Isard}, Y.~{Jia},
  R.~{Jozefowicz}, L.~{Kaiser}, M.~{Kudlur}, J.~{Levenberg}, D.~{Mane},
  R.~{Monga}, S.~{Moore}, D.~{Murray}, C.~{Olah}, M.~{Schuster}, J.~{Shlens},
  B.~{Steiner}, I.~{Sutskever}, K.~{Talwar}, P.~{Tucker}, V.~{Vanhoucke},
  V.~{Vasudevan}, F.~{Viegas}, O.~{Vinyals}, P.~{Warden}, M.~{Wattenberg},
  M.~{Wicke}, Y.~{Yu}, and X.~{Zheng}.
\newblock {TensorFlow: Large-Scale Machine Learning on Heterogeneous
  Distributed Systems}.
\newblock Mar. 2016.

\bibitem{2017arXiv170201105A}
I.~Armeni, S.~Sax, A.~R. Zamir, and S.~Savarese.
\newblock Joint 2d-3d-semantic data for indoor scene understanding.
\newblock {\em CoRR}, abs/1702.01105, 2017.

\bibitem{atapour2018real}
A.~Atapour-Abarghouei and T.~P. Breckon.
\newblock Real-time monocular depth estimation using synthetic data with domain
  adaptation via image style transfer.
\newblock In {\em Proceedings of the IEEE Conference on Computer Vision and
  Pattern Recognition}, volume~18, page~1, 2018.

\bibitem{bousmalis2017unsupervised}
K.~Bousmalis, N.~Silberman, D.~Dohan, D.~Erhan, and D.~Krishnan.
\newblock Unsupervised pixel-level domain adaptation with generative
  adversarial networks.
\newblock In {\em The IEEE Conference on Computer Vision and Pattern
  Recognition (CVPR)}, volume~1, page~7, 2017.

\bibitem{Matterport3D}
A.~Chang, A.~Dai, T.~Funkhouser, M.~Halber, M.~Niessner, M.~Savva, S.~Song,
  A.~Zeng, and Y.~Zhang.
\newblock {Matterport3D}: Learning from {RGB-D} data in indoor environments.
\newblock {\em International Conference on 3D Vision (3DV)}, 2017.

\bibitem{deeplab}
L.-C. Chen, G.~Papandreou, I.~Kokkinos, K.~Murphy, and A.~L.~Yuille.
\newblock Deeplab: Semantic image segmentation with deep convolutional nets,
  atrous convolution, and fully connected crfs.
\newblock In {\em IEEE Transactions on Pattern Analysis and Machine
  Intelligence}, volume~PP, 06 2016.

\bibitem{chen2017no}
Y.-H. Chen, W.-Y. Chen, Y.-T. Chen, B.-C. Tsai, Y.-C.~F. Wang, and M.~Sun.
\newblock No more discrimination: Cross city adaptation of road scene
  segmenters.
\newblock In {\em Computer Vision (ICCV), 2017 IEEE International Conference
  on}, pages 2011--2020. IEEE, 2017.

\bibitem{cityscapes}
M.~Cordts, M.~Omran, S.~Ramos, T.~Rehfeld, M.~Enzweiler, R.~Benenson,
  U.~Franke, S.~Roth, and B.~Schiele.
\newblock The cityscapes dataset for semantic urban scene understanding.
\newblock In {\em 2016 IEEE Conference on Computer Vision and Pattern
  Recognition (CVPR)}, pages 3213--3223, 2016.

\bibitem{Augmentation19}
T.~Dao, A.~Gu, A.~Ratner, V.~Smith, C.~D. Sa, and C.~R{\'{e}}.
\newblock A kernel theory of modern data augmentation.
\newblock In {\em Proceedings of the 36th International Conference on Machine
  Learning, {ICML} 2019, 9-15 June 2019, Long Beach, California, {USA}}, pages
  1528--1537, 2019.

\bibitem{imagenet}
J.~Deng, W.~Dong, R.~Socher, L.~jia Li, K.~Li, and L.~Fei-fei.
\newblock Imagenet: A large-scale hierarchical image database.
\newblock In {\em In CVPR}, 2009.

\bibitem{gaidon2016virtual}
A.~Gaidon, Q.~Wang, Y.~Cabon, and E.~Vig.
\newblock Virtual worlds as proxy for multi-object tracking analysis.
\newblock In {\em Proceedings of the IEEE conference on computer vision and
  pattern recognition}, pages 4340--4349, 2016.

\bibitem{ganin2015unsupervised}
Y.~Ganin and V.~Lempitsky.
\newblock Unsupervised domain adaptation by backpropagation.
\newblock In {\em Proceedings of the 32nd International Conference on
  International Conference on Machine Learning-Volume 37}, pages 1180--1189.
  JMLR. org, 2015.

\bibitem{GaninJournal}
Y.~Ganin, E.~Ustinova, H.~Ajakan, P.~Germain, H.~Larochelle, F.~Laviolette,
  M.~Marchand, and V.~Lempitsky.
\newblock Domain-adversarial training of neural networks.
\newblock {\em J. Mach. Learn. Res.}, 17(1):2096--2030, Jan. 2016.

\bibitem{Gatys}
L.~A. Gatys, A.~S. Ecker, and M.~Bethge.
\newblock A neural algorithm of artistic style.
\newblock volume abs/1508.06576, 2015.

\bibitem{goodfellow2014generative}
I.~Goodfellow, J.~Pouget-Abadie, M.~Mirza, B.~Xu, D.~Warde-Farley, S.~Ozair,
  A.~Courville, and Y.~Bengio.
\newblock Generative adversarial nets.
\newblock In {\em Advances in neural information processing systems}, pages
  2672--2680, 2014.

\bibitem{resnet101}
K.~He, X.~Zhang, S.~Ren, and J.~Sun.
\newblock Deep residual learning for image recognition.
\newblock In {\em 2016 {IEEE} Conference on Computer Vision and Pattern
  Recognition, {CVPR} 2016, Las Vegas, NV, USA, June 27-30, 2016}, pages
  770--778, 2016.

\bibitem{CyCADA}
J.~Hoffman, E.~Tzeng, T.~Park, J.-Y. Zhu, P.~Isola, K.~Saenko, A.~A. Efros, and
  T.~Darrell.
\newblock Cycada: Cycle-consistent adversarial domain adaptation.
\newblock In {\em ICML}, 2018.

\bibitem{inthewild}
J.~Hoffman, D.~Wang, F.~Yu, and T.~Darrell.
\newblock Fcns in the wild: Pixel-level adversarial and constraint-based
  adaptation.
\newblock In {\em CoRR}, volume abs/1612.02649, 2016.

\bibitem{densenet}
G.~Huang, Z.~Liu, L.~van~der Maaten, and K.~Q. Weinberger.
\newblock Densely connected convolutional networks.
\newblock {\em 2017 IEEE Conference on Computer Vision and Pattern Recognition
  (CVPR)}, pages 2261--2269, 2017.

\bibitem{huang2016deep}
G.~Huang, Y.~Sun, Z.~Liu, D.~Sedra, and K.~Q. Weinberger.
\newblock Deep networks with stochastic depth.
\newblock In {\em European Conference on Computer Vision}, pages 646--661.
  Springer, 2016.

\bibitem{apollo}
X.~Huang, X.~Cheng, Q.~Geng, B.~Cao, D.~Zhou, P.~Wang, Y.~Lin, and R.~Yang.
\newblock The {ApolloScape} dataset for autonomous driving.
\newblock In {\em 2018 {IEEE}/{CVF} Conference on Computer Vision and Pattern
  Recognition Workshops ({CVPRW})}. {IEEE}, jun 2018.

\bibitem{patchGAN}
P.~Isola, J.-Y. Zhu, T.~Zhou, and A.~A. Efros.
\newblock Image-to-image translation with conditional adversarial networks.
\newblock In {\em 2017 IEEE Conference on Computer Vision and Pattern
  Recognition (CVPR)}, pages 5967--5976, 2017.

\bibitem{ArtisticVia}
P.~T. Jackson, A.~Atapour-Abarghouei, S.~Bonner, T.~P. Breckon, and B.~Obara.
\newblock Style augmentation: Data augmentation via style randomization.
\newblock In {\em The IEEE Conference on Computer Vision and Pattern
  Recognition (CVPR) Workshops}, June 2019.

\bibitem{MNIST}
Y.~Lecun, L.~Bottou, Y.~Bengio, and P.~Haffner.
\newblock Gradient-based learning applied to document recognition.
\newblock In {\em Proceedings of the IEEE}, pages 2278--2324, 1998.

\bibitem{semAwareGradGAN}
P.~Li, X.~Liang, D.~Jia, and E.~Xing.
\newblock Semantic-aware grad-gan for virtual-to-real urban scene adaption.
\newblock In {\em British Machine Vision Conference 2018, {BMVC} 2018,
  Northumbria University, Newcastle, UK, September 3-6, 2018}, page~73, 2018.

\bibitem{fastphoto}
Y.~Li, M.~Liu, X.~Li, M.~Yang, and J.~Kautz.
\newblock A closed-form solution to photorealistic image stylization.
\newblock In {\em Computer Vision - {ECCV} 2018 - 15th European Conference,
  Munich, Germany, September 8-14, 2018, Proceedings, Part {III}}, pages
  468--483, 2018.

\bibitem{liu2016coupled}
M.-Y. Liu and O.~Tuzel.
\newblock Coupled generative adversarial networks.
\newblock In {\em NIPS}, 2016.

\bibitem{FCN}
J.~Long, E.~Shelhamer, and T.~Darrell.
\newblock Fully convolutional networks for semantic segmentation.
\newblock In {\em {IEEE} Conference on Computer Vision and Pattern Recognition,
  {CVPR} 2015, Boston, MA, USA, June 7-12, 2015}, pages 3431--3440, 2015.

\bibitem{menze2015object}
M.~Menze and A.~Geiger.
\newblock Object scene flow for autonomous vehicles.
\newblock In {\em Proceedings of the IEEE Conference on Computer Vision and
  Pattern Recognition}, pages 3061--3070, 2015.

\bibitem{im2im2}
Z.~Murez, S.~Kolouri, D.~J. Kriegman, R.~Ramamoorthi, and K.~Kim.
\newblock Image to image translation for domain adaptation.
\newblock In {\em 2018 {IEEE} Conference on Computer Vision and Pattern
  Recognition, {CVPR} 2018, Salt Lake City, UT, USA, June 18-22, 2018}, pages
  4500--4509, 2018.

\bibitem{mapillary}
G.~Neuhold, T.~Ollmann, S.~Rota~Bulo, and P.~Kontschieder.
\newblock The mapillary vistas dataset for semantic understanding of street
  scenes.
\newblock In {\em The IEEE International Conference on Computer Vision (ICCV)},
  Oct 2017.

\bibitem{pytorch}
A.~Paszke, S.~Gross, S.~Chintala, G.~Chanan, E.~Yang, Z.~DeVito, Z.~Lin,
  A.~Desmaison, L.~Antiga, and A.~Lerer.
\newblock Automatic differentiation in pytorch.
\newblock 2017.

\bibitem{GTA}
S.~R. Richter, V.~Vineet, S.~Roth, and V.~Koltun.
\newblock Playing for data: Ground truth from computer games.
\newblock In {\em ECCV}, 2016.

\bibitem{romijnders2018domain}
R.~Romijnders, P.~Meletis, and G.~Dubbelman.
\newblock A domain agnostic normalization layer for unsupervised adversarial
  domain adaptation.
\newblock In {\em {IEEE} Winter Conference on Applications of Computer Vision,
  {WACV} 2019, Waikoloa Village, HI, USA, January 7-11, 2019}, pages
  1866--1875, 2019.

\bibitem{SYNTHIA}
G.~Ros, L.~Sellart, J.~Materzynska, D.~Vazquez, and A.~M. Lopez.
\newblock The synthia dataset: A large collection of synthetic images for
  semantic segmentation of urban scenes.
\newblock In {\em The IEEE Conference on Computer Vision and Pattern
  Recognition (CVPR)}, June 2016.

\bibitem{saenko2010adapting}
K.~Saenko, B.~Kulis, M.~Fritz, and T.~Darrell.
\newblock Adapting visual category models to new domains.
\newblock In {\em European conference on computer vision}, pages 213--226.
  Springer, 2010.

\bibitem{dropoutRegularization}
K.~Saito, Y.~Ushiku, T.~Harada, and K.~Saenko.
\newblock Adversarial dropout regularization.
\newblock In {\em International Conference on Learning Representations}, 2018.

\bibitem{sankaranarayanan2018learning}
S.~Sankaranarayanan, Y.~Balaji, A.~Jain, S.~N. Lim, and R.~Chellappa.
\newblock Learning from synthetic data: Addressing domain shift for semantic
  segmentation.
\newblock In {\em The IEEE Conference on Computer Vision and Pattern
  Recognition (CVPR)}, 2018.

\bibitem{VGG}
K.~Simonyan and A.~Zisserman.
\newblock Very deep convolutional networks for large-scale image recognition.
\newblock In {\em International Conference on Learning Representations}, 2015.

\bibitem{song2016ssc}
S.~Song, F.~Yu, A.~Zeng, A.~X. Chang, M.~Savva, and T.~Funkhouser.
\newblock Semantic scene completion from a single depth image.
\newblock {\em IEEE Conference on Computer Vision and Pattern Recognition},
  2017.

\bibitem{srivastava2014dropout}
N.~Srivastava, G.~Hinton, A.~Krizhevsky, I.~Sutskever, and R.~Salakhutdinov.
\newblock Dropout: a simple way to prevent neural networks from overfitting.
\newblock {\em The Journal of Machine Learning Research}, 15(1):1929--1958,
  2014.

\bibitem{stein2018genesis}
G.~J. Stein and N.~Roy.
\newblock Genesis-rt: Generating synthetic images for training secondary
  real-world tasks.
\newblock In {\em 2018 IEEE International Conference on Robotics and Automation
  (ICRA)}, pages 7151--7158. IEEE, 2018.

\bibitem{ArtisticBy}
C.~Thomas and A.~Kovashka.
\newblock Artistic object recognition by unsupervised style adaptation.
\newblock In C.~V. Jawahar, H.~Li, G.~Mori, and K.~Schindler, editors, {\em
  Computer Vision -- ACCV 2018}, pages 460--476, Cham, 2019. Springer
  International Publishing.

\bibitem{torralba2011unbiased}
A.~Torralba and A.~A. Efros.
\newblock Unbiased look at dataset bias.
\newblock In {\em Computer Vision and Pattern Recognition (CVPR), 2011 IEEE
  Conference on}, pages 1521--1528. IEEE, 2011.

\bibitem{Tsai_2018_CVPR}
Y.-H. Tsai, W.-C. Hung, S.~Schulter, K.~Sohn, M.-H. Yang, and M.~Chandraker.
\newblock Learning to adapt structured output space for semantic segmentation.
\newblock In {\em The IEEE Conference on Computer Vision and Pattern
  Recognition (CVPR)}, June 2018.

\bibitem{phashing}
X.~Wang, K.~Pang, X.~Zhou, Y.~Zhou, L.~Li, and J.~Xue.
\newblock A visual model-based perceptual image hash for content
  authentication.
\newblock volume~10, pages 1336--1349, 2015.

\bibitem{resnet38}
Z.~Wu, C.~Shen, and A.~van~den Hengel.
\newblock Wider or deeper: Revisiting the resnet model for visual recognition.
\newblock In {\em CoRR}, volume abs/1611.10080, 2019.

\bibitem{xiao2012recognizing}
J.~Xiao, K.~A. Ehinger, A.~Oliva, and A.~Torralba.
\newblock Recognizing scene viewpoint using panoramic place representation.
\newblock In {\em Computer Vision and Pattern Recognition (CVPR), 2012 IEEE
  Conference on}, pages 2695--2702. IEEE, 2012.

\bibitem{yosinski2014transferable}
J.~Yosinski, J.~Clune, Y.~Bengio, and H.~Lipson.
\newblock How transferable are features in deep neural networks?
\newblock In {\em Advances in neural information processing systems}, pages
  3320--3328, 2014.

\bibitem{zhang2017curriculum}
Y.~Zhang, P.~David, and B.~Gong.
\newblock Curriculum domain adaptation for semantic segmentation of urban
  scenes.
\newblock In {\em The IEEE International Conference on Computer Vision (ICCV)},
  volume~2, page~6, 2017.

\bibitem{im2imCycle}
J.-Y. Zhu, T.~Park, P.~Isola, and A.~A. Efros.
\newblock Unpaired image-to-image translation using cycle-consistent
  adversarial networks.
\newblock In {\em Computer Vision (ICCV), 2017 IEEE International Conference
  on}, 2017.

\bibitem{Zioulis_2018_ECCV}
N.~Zioulis, A.~Karakottas, D.~Zarpalas, and P.~Daras.
\newblock Omnidepth: Dense depth estimation for indoors spherical panoramas.
\newblock In {\em The European Conference on Computer Vision (ECCV)}, September
  2018.

\bibitem{DAclassBalanced}
Y.~Zou, Z.~Yu, B.~Vijaya~Kumar, and J.~Wang.
\newblock Unsupervised domain adaptation for semantic segmentation via
  class-balanced self-training.
\newblock In {\em The European Conference on Computer Vision (ECCV)}, September
  2018.

\end{thebibliography}
}

\end{document}